\documentclass[conference]{IEEEtran}

\usepackage{spconf,amsmath,graphicx}
\usepackage{amsfonts}
\usepackage{booktabs}
\usepackage{multirow}
\usepackage{hyperref}
\usepackage{xcolor}
\usepackage{soul}
\usepackage{float}
\usepackage{graphicx}  
\usepackage{verbatim}
\usepackage{tablefootnote}
\usepackage{threeparttable} 
\usepackage{bm}
\usepackage[mathscr]{euscript}

\usepackage{array}

\usepackage[%
    format=hang,    
    format=plain,
    margin=0pt,
]{caption}
\usepackage{subcaption}

\title{PointPCA+: Extending PointPCA objective quality assessment metric}
\name{Xuemei Zhou\textsuperscript{1,3},
      Evangelos Alexiou\textsuperscript{2},
      Irene Viola\textsuperscript{1},
      Pablo Cesar\textsuperscript{1,3}}
\address{\textsuperscript{1}Centrum Wiskunde $\&$ Informatica, Amsterdam, The Netherlands,\\
\textsuperscript{2}TNO Netherlands Organisation for Applied Scientific Research, The Hague, The Netherlands,\\
\textsuperscript{3}TU Delft, Delft, The Netherlands
}

\begin{document}
%
\maketitle
\begin{abstract}
A computationally-simplified and descriptor-richer Point Cloud Quality Assessment (PCQA) metric, namely PointPCA+, is proposed in this paper, which is an extension of PointPCA. 
PointPCA proposed a set of perceptually-relevant descriptors based on PCA decomposition that were applied to both the geometry and texture data of point clouds for full reference PCQA. 
PointPCA+ employs PCA only on the geometry data while enriching existing geometry and texture descriptors, that are computed more efficiently. 
Similarly to PointPCA, a total quality score is obtained through a learning-based fusion of individual predictions from geometry and texture descriptors that capture local shape and appearance properties, respectively.
Before feature fusion, a feature selection module is introduced to choose the most effective features from a proposed super-set.
Experimental results show that PointPCA+ achieves high predictive performance against subjective ground truth scores obtained from publicly available datasets. 
The code is available at \url{https://github.com/cwi-dis/pointpca_suite/}.
\end{abstract}
\begin{keywords}
Point cloud, PCA, Feature selection, Objective quality assessment, Random forest
\end{keywords}

\section{Introduction}
\label{sec:intro}
Point cloud is prevailing among the available 3D imaging formats in recent years. It is essentially a collection of points, where each point has attributes of geometry, color, reflectance, etc. However, through acquisition, compression, transmission, and rendering, the quality of a point cloud can be degraded, which necessitates effective and efficient Point Cloud Quality Assessment (PCQA) metrics. These metrics provide a guide on the design, optimization, and parameter tuning of point cloud processing pipelines.

Objective PCQA metrics can be divided into point-based, projection-based, and feature-based models. Point-based metrics such as point-to-point, point-to-plane metrics and their variants measure degradations between the original and distorted point clouds per point, mainly based on Euclidean or color space distances~\cite{ALEXIOU2023501}. Alexiou \textit{et al.} propose the angular similarity of tangent planes among corresponding points, which considers neighborhood information~\cite{8486512}. These metrics are computationally efficient but suffer from a crude correspondence of matching between points.

\begin{figure*}[htbp]
\centering
\includegraphics[width=0.9\textwidth]{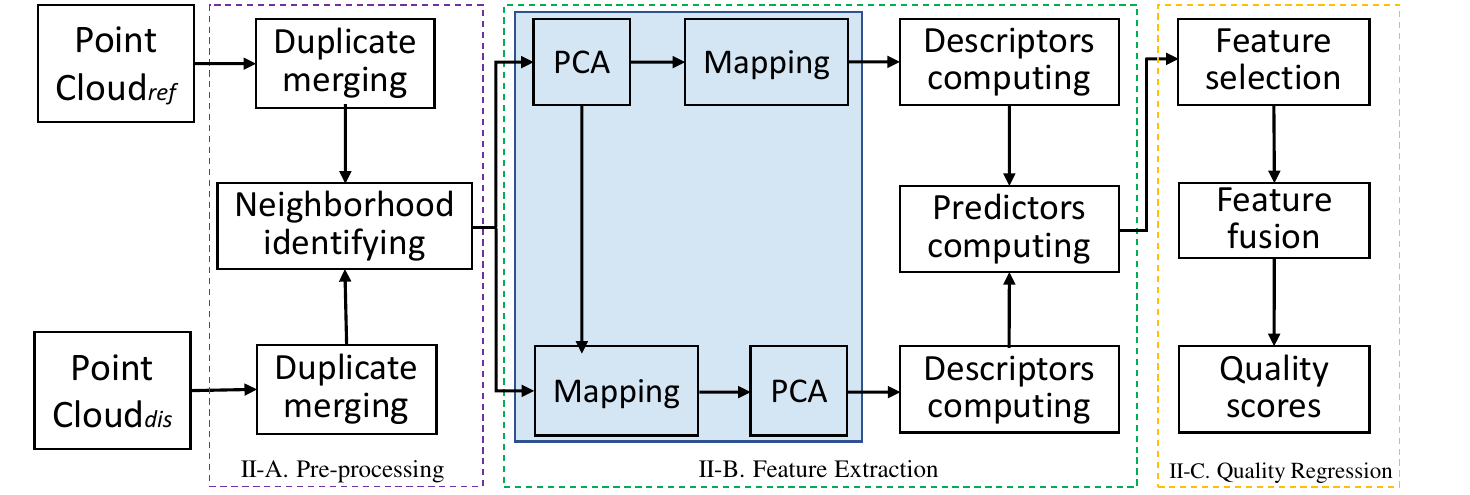}
\caption{PointPCA+ architecture: both the reference and the distorted point cloud are passing from every stage to compute a quality score. Operations in the blue box are applied only to the geometry data of point clouds. 
} 
\label{fig:architecture}

\vspace{-0.5em}
\end{figure*}
Projection-based approaches adapt existing Image Quality Assessment (IQA) metrics to PCQA. Alexiou \textit{et al.}~\cite{8743277} develop a framework for predicting the quality by employing 2D IQA metrics on 6 orthographic projected views. 
Liu \textit{et al.}~\cite{9756929} provide a PCQA model based on the principle of information content weighted structural similarity (IW-SSIM). However, the projection process and the number of viewpoints have a non-negligible impact on the final prediction accuracy; besides, how to combine the quality score on each viewpoint into a score is also not straightforward.

Feature-based metrics consider perceptual loss from both geometry and texture properties. Viola \textit{et al.}~\cite{9123089} extract color statistics, histogram, and correlogram to assess the level of impairment and combine the color-based metrics with geometry-based metrics to form a global quality score. Alexiou \textit{et al.}~\cite{9106005} employ the local distributions of point clouds to predict perceptual degradations from topology and color. Yang \textit{et al.}~\cite{9306905} construct a local graph centered at resampled key points for both reference and distorted point clouds, with the color gradient on the local graph being used to measure distortions. 
Meynet \textit{et al.}~\cite{9123147} utilize an optimally-weighted linear combination of curvature-based and color-based features to evaluate visual quality. Diniz \textit{et al.}~\cite{diniz2022point} adopt the statistical information of the extracted geometry/color features and feed them into a regression model. 
Deep learning-based modules have also been used to extract perceptual features. Liu \textit{et al.}~\cite{liu2023point} design an end-to-end No Reference (NR) PCQA framework for estimating subjective quality. A sparse convolutional neural network is applied in both feature extraction and regression modules. 
An extension using coarse-to-fine progressive knowledge transfer based on Human Vision System (HVS) is given in~\cite{liu2022progressive}.   
Zhang \textit{et al.} \cite{zhang2022mm} make use of multi-modal information to address the PCQA problem; the quality-aware encoder features are optimized with the assistance of symmetric cross-modality attention. Interested readers may refer to  \cite{ALEXIOU2023501} for a more comprehensive review of the literature.

Point-based schemes may neglect the high-dimensional properties of point clouds and the interplay among these dimensions, thereby limiting their effectiveness. Projection-based methods often rely on 2D IQA, which may not adequately capture the intrinsic characteristics of point clouds. 
Feature-based schemes tend to have a high level of complexity, while the interpretability of deep learning-based methods is a drawback with training requiring a huge amount of data. In light of the above limitations and inspired by PointPCA \cite{alexiou2021pointpca}, we propose PointPCA+. Our differences are threefold compared with PointPCA.
\begin{itemize}
    \item By performing PCA on the geometry data of the reference point cloud and transforming both the reference and distorted point clouds onto the new basis, we can capture differences in their shape properties effectively.
    \item We utilize 
    $knn$ algorithm to determine the neighborhood, which is faster and returns a consistent number of points, therefore further decreasing the computational cost of subsequent processing steps. 
    \item Recursive Feature Elimination (RFE) algorithm is employed to select the most relevant and complementary features. This approach streamlines the quality regression process. 
\end{itemize}

\section{Proposed PointPCA+ Method}
In Fig.~\ref{fig:architecture}, the PointPCA+ framework is illustrated, which is split into three modules, namely, (a) pre-processing, (b) feature extraction, and (c) quality regression which are introduced in the following subsections. Note that a Full Reference (FR) metric typically uses either the pristine or the impaired content as a reference, or both. In our design, only the pristine point cloud serves as a reference.

\subsection{Pre-processing}
To ensure coherent geometry and color information without redundancies, points with identical coordinates that belong to the same point cloud are merged \cite{8296925}. The color of a merged point is obtained by averaging the color of corresponding points sharing the same coordinates. 
For an FR PCQA metric, identifying matches between reference and distorted point clouds is crucial for comparing corresponding local properties.
In our method, we use the $knn$ algorithm to identify neighborhood pairs between two point clouds. In particular, for each point that belongs to a reference point cloud $\mathcal{A}$, we find its $K$ nearest reference points, and its $K$ nearest distorted points from the distorted point cloud $\mathcal{B}$, in terms of Euclidean distance. 

\subsection{Feature Extraction}
To capture local perceptual quality degradations of a distorted point cloud, we compute geometry and texture descriptors based on the identified neighborhoods. Statistics based on these descriptors are subsequently calculated and serve as predictors of visual quality. 
As mentioned earlier, our method uses only the pristine point cloud as a reference to find the matches in the distorted point cloud.

\subsubsection*{Geometry descriptors}
Given a query point $\mathbf{p}_i$ of $\mathcal{A}$, the subscript $i$ denotes the point index, $1 \leq i \leq |\mathcal{A}|$, and $|\mathcal{A}|$ is the cardinality. 
The coordinates of $\mathbf{p}_i$'s $N$ nearest neighbors in $\mathcal{F}$ are indicated as 
$\bm{p}^{g,\mathcal{F}}_{n} = ( x_n, y_n, z_n )^{\rm T}$, with ${1 \le n \le N}$ and $\mathcal{F}\in\{{\mathcal{A},\mathcal{B}}\}$.
The geometry of $\mathbf{p}_i$ 
is denoted as $\bm{p}_{i}^{g, \mathcal{A}}$, and the geometry of its closest neighbor in $\mathcal{B}$ is denoted as $\bm{p}_{i}^{g, \mathcal{B}}$.

Initially, the covariance matrix $\bm \Sigma_{i}^{\mathcal{A}}$ 
is computed as

\begin{equation}
\bm{\Sigma}_i^{\mathcal{A}} = \frac{1}{N} \textstyle \sum_{n = 1}^{N} 
\left(\bm{p}^{g,\mathcal{A}}_n - \bar{\bm{{p}}}^{g,\mathcal{A}}_{i}\right) \cdot \left(\bm{p}^{g,\mathcal{A}}_n - \bar{\bm{{p}}}^{g,\mathcal{A}}_{i}\right)^{\rm T},
\label{eq:cov}
\end{equation}
where $\bar{\bm{p}}^{g,\mathcal{A}}_{i} $ indicates the centroid, given as
\begin{equation}
\bm{\bar{{p}}}^{g,\mathcal{A}}_i = \frac{1}{N} \textstyle \sum_{n = 1}^{N} 
\bm{p}^{g,\mathcal{A}}_n.
\label{eq:centroid}
\end{equation}

\begin{table}[tbp]
\caption{Definition of descriptors.}
\vspace{-0.5em}
\centering
\renewcommand{\arraystretch}{1.6}
\resizebox{1\linewidth}{!}{
\begin{tabular}{l l l l}
\toprule
& Descriptor          & Definition   & Distance     \\ \midrule
\multirow{16}{*}{\rotatebox[origin=c]{90}{{\centering Geometric}}}  
& Error vector   & $\bm{e} = (\bm{\omega}_{i}^{\mathcal{B}}-\bm{\omega}_{i}^{\mathcal{A}})$ & $r_{\alpha}$\\
& Error along axes    & $\epsilon_{m} = (\bm{\omega}_{i}^{\mathcal{B}}-\bm{\omega}_{i}^{\mathcal{A}}) ^ {T}\cdot \mathbf{u}_{m}$ &$r_{\beta}$\\
& Error from origin  & $\bm{\mathcal{\varepsilon}} = \bm{\omega}_{i}^{\mathcal{F}}$ & $r_{\alpha}$,$r_{\beta}$\\
& Mean          & $\bm{\mu}^{\mathcal{B}} = \frac{1}{N}\textstyle \sum_{n}\bm{\omega}_{n}^{\mathcal{B}}$ &$r_{\alpha}$,$r_{\beta}$\\
& Variance     & $\bm{\lambda}^{\mathcal{F}} = \frac{1}{N} \textstyle \sum_{n}{\left(\bm{\omega}_n^{\mathcal{F}} - \bm{\mu}^{\mathcal{F}}\right)}^2$ &$r_{\delta}$\\
& Sum of variance           & $\Sigma^{\mathcal{F}} = \textstyle \sum_{m}\lambda_{m}^{\mathcal{F}}$ &$r_{\delta}$\\
& Covariance                & \tiny{$\bm{\Sigma} = \frac{1}{N} \textstyle \sum_{n}\left(\bm{\omega}_n^{\mathcal{A}} - \bm{\mu}^{\mathcal{A}}\right) \cdot \left(\bm{\omega}_n^{\mathcal{B}} - \bm{\mu}^{\mathcal{B}}\right)^{\rm T}$} &$r_{\gamma}$\\
& Omnivariance              & ${\rm O}^{\mathcal{F}} = \sqrt[3]{\textstyle \prod_{m}\lambda_{m}^{\mathcal{F}}}$ &$r_{\lambda}$\\
& Eigenentropy             & ${\rm E}^{ \mathcal{F}} = -\textstyle \sum_{m} {\lambda}^{\mathcal{F}}_{m} \cdot \log{\lambda^{\mathcal{F}}_{m}}$ &$r_{\delta}$\\
& Anisotropy             & $ {\rm A}^{\mathcal{F}} = ({\lambda}^{\mathcal{F}}_{1} - {\lambda}^{\mathcal{F}}_{3} ) / {\lambda}^{\mathcal{F}}_{1}$ &$r_{\delta}$\\
& Planarity              & ${\rm P}^{\mathcal{F}} = ({\lambda}^{\mathcal{F}}_{2}  - {\lambda}^{\mathcal{F}}_{3} ) / {\lambda}^{\mathcal{F}}_{1}$ &$r_{\delta}$\\ 
& Linearity              & ${\rm L}^{\mathcal{F}} = ({\lambda}^{\mathcal{F}}_{1} - {\lambda}^{\mathcal{F}}_{2})/ {\lambda}^{\mathcal{F}}_{1}$ &$r_{\delta}$\\
& Scattering             & ${\rm S}^{\mathcal{F}} = {\lambda}^{\mathcal{F}}_{3} / {\lambda}^{\mathcal{F}}_{1}$ &$r_{\delta}$\\ 
& Change of curvature      & ${\rm C}^{\mathcal{F}} ={\lambda}^{\mathcal{F}}_{3} \big/ \textstyle \sum_{m} {\lambda}^{\mathcal{F}}_{m}$ &$r_{\delta}$\\
& Parallelity           & $\mathscr{P}_{m} = 1 - \mathbf{u}_{m} \cdot {\mathbf{v}^{\mathcal B}_{m} }$ &$-$\\
& Angular similarity    & $\theta_{m} = 1- \frac {2 \cdot \arccos(\cos(\mathbf{u}_{m},\mathbf{v}^{\mathcal B}_{m}))} {\pi}$ &$-$\\
\midrule
\multirow{6}{*}{\rotatebox[origin=c]{90}{{\centering Textural}}}
& Mean             & $\tilde{\bm{\mu}}^{\mathcal{F}} =  \frac{1}{N}\textstyle \sum_{n} \bm{p}^{t,\mathcal{F}}_{n}$ &$r_{\delta}$\\
& Variance         & $\tilde{\bm{s}}^{\mathcal{F}} =  \frac{1}{N} \textstyle \sum_{n} \left(\bm{p}^{t,\mathcal{F}}_{n} - \tilde{\bm{\mu}}^{\mathcal{F}}\right)^2$ & $r_{\delta}$\\
& Sum of variance  & $\tilde{\Sigma}^{\mathcal{F}}= \textstyle\sum_{m}\tilde{s}^{\mathcal{F}}_{m}$ & $r_{\delta}$\\
& Covariance       & \tiny{$\tilde{\bm{\Sigma}} = \frac{1}{N} \textstyle \sum_{n}\left(\bm{p}_n^{t,\mathcal{A}} - \tilde{\bm{\mu}}^{\mathcal{A}}\right) \cdot \left(\bm{p}_n^{t,\mathcal{B}} - \tilde{\bm{\mu}}^{\mathcal{B}}\right)^{\rm T}$} &$r_{\gamma}$\\
& Omnivariance     & $\tilde{\rm{O}}^{\mathcal{F}} = \sqrt[3]{\textstyle \prod_{m}{\tilde{s}^{\mathcal{F}}_{m}}}$  &$r_{\delta}$\\
& Entropy          & $\tilde{\rm{H}}^{\mathcal{F}} = - \textstyle \sum_{m} \tilde{s}^{\mathcal{F}}_{m} \cdot \log{\tilde{s}^{\mathcal{F}}_{m}}$ & $r_{\delta}$ \\
\bottomrule
\end{tabular}}
\label{tbl:descript}
\end{table}

\noindent Then, eigen-decomposition is  applied to $\bm{\Sigma}_i^{\mathcal{A}}$, to obtain the eigenvectors which form an orthonormal basis $\mathbf{V}^\mathcal{A}$ composed of eigenvectors $\mathbf{v}_{m}^\mathcal{A}$, with corresponding eigenvalues $\lambda_m^{\mathcal{A}}$, where $m = {1,2,3}$, and $\lambda_1^{\mathcal{A}} > \lambda_2^{\mathcal{A}} > \lambda_3^{\mathcal{A}}$. 
Next, we map the reference and distorted neighborhoods to the new orthonormal basis, denoted as  
$\bm{\omega}_{n}^{\mathcal{F}} = (\bm{p}^{g,\mathcal{F}}_{n}-\bar{\bm{{p}}}^{g\,\mathcal{A}}_{i})\cdot \mathbf{V}^{\mathcal A}$. 
Finally, we apply PCA to the covariance matrix of $\bm{\omega}_{n}^{\mathcal{B}}$ and compute the eigenvectors $\mathbf{v}_{m}^ \mathcal B$ and eigenvalues $\lambda_{m}^{\mathcal B}$. 
The mapped coordinates of the reference and distorted points 
$\bm{\omega}_n^{\mathcal{F}}$,
the eigenvectors $\mathbf{v}_{m}^\mathcal{F}$ 
and the unit vectors $\mathbf{u}_{m}$, with $\mathbf{u}_{1} = [1,0,0]^{\rm{T}}$, $\mathbf{u}_{2} = [0,1,0]^{\rm{T}}$ and $\mathbf{u}_{3} = [0,0,1]^{\rm{T}}$, are used to construct the geometric descriptors defined in Table~\ref{tbl:descript}.

\subsubsection*{Texture descriptors}
The color space is first converted from RGB to YCbCr \cite{ITURBT7096}. This conversion is motivated by the fact that the human eye is more sensitive to changes in brightness than changes in color. 
We denote the texture information of $\mathbf{p}_i$'s $N$ nearest neighbors in $\mathcal{F}$ as $\bm{p}^{t,\mathcal{F}}_{n} = ( Y_n, Cb_n, Cr_n)^\mathrm{T}$. The proposed 6 texture descriptors are defined in Table \ref{tbl:descript}.

\subsubsection*{Explanation of descriptors}
Each geometry descriptor represents an interpretable shape property inside the neighborhood. 
Specifically, 
$\bm{e}$ denotes the error vector between the mapped coordinates of the reference query point and its nearest neighbor, and $\epsilon_{m}$ is the projected distance of the error vector across the $m$-th axis. 
The $\bm{\mathcal{\varepsilon}}$ is used to capture the Euclidean and projected distances of the mapped reference query point or its nearest distorted neighbor from the centroid and principal axes, respectively.
$\bm{\mu}^{\mathcal{B}}$, $\bm {\lambda}^{\mathcal{F}}$, $\Sigma^{\mathcal{F}}$ and $\bm{\Sigma}$ reveal local statistics. $\rm{E}^{\mathcal{F}}$ provides an estimation of the space uncertainty on the projected surfaces. Additionally, $\mathscr{P}_{m}$ and $\theta_{m}$ assess the parallelity and the angular dispersion of the distorted plane. 
The remaining geometry descriptors explore the topology of a local region from different aspects, relying on the spatial dispersion along different principal axes. $\tilde{\bm{\mu}}^{\mathcal{F}}$, $\tilde{\bm{s}}^{\mathcal{F}}$ and $\tilde{\Sigma}^{\mathcal{F}}$ of the YCbCr channel express the intrinsic distribution of luminance and chromatic components. $\tilde{\bm{\Sigma}}$ and $\tilde{\rm{O}}^{\mathcal{F}}$ show the variability of color information. $\tilde{\rm{\bm{H}}}^{\mathcal{F}}$ provides an estimation of color uncertainty of the local region.
Every descriptor is computed per point $\mathbf{p}_i$.

\begin{table*}[]
\caption{ SROCC performance  on M-PCCD, SJTU and WPC datasets}
\centering
\vspace{-0.5em}
\renewcommand{\arraystretch}{1.1}
\scalebox{0.89}{
\begin{tabular}{c|c|c|c|c|c|c|c|c|c}
\hline
Metric & PointPCA+   & PointPCA\cite{alexiou2021pointpca}    & PCQM\cite{9123147}        & pSSIM\cite{9106005} & BitDance\cite{9450013}    & Plane2plane\cite{8486512}  & P2Plane\_MSE\cite{ALEXIOU2023501} & P2P\_MSE \cite{ALEXIOU2023501} & PSNR Y\cite{ALEXIOU2023501}     \\ \hline
M-PCCD  & \textbf{0.943}$\pm$0.022 & \underline{0.941}$\pm$0.032 & 0.940$\pm$0.032 & 0.925$\pm$0.024  & 0.859$\pm$0.061 & 0.847$\pm$0.076& 0.901$\pm$0.025  & 0.896$\pm$0.042 & 0.798$\pm$0.162 \\
SJTU   & \underline{0.865}$\pm$0.064 & \textbf{0.890}$\pm$0.056 & 0.862$\pm$0.030 & 0.708$\pm$0.070  & 0.748$\pm$0.077 & 0.761$\pm$0.039 & 0.578$\pm$0.155  & 0.612$\pm$0.157 & 0.743$\pm$0.083 \\
WPC    & \underline{0.857}$\pm$0.040 & \textbf{0.866}$\pm$0.036 & 0.749$\pm$0.036 & 0.465$\pm$0.059  & 0.451$\pm$0.054 & 0.454$\pm$0.069 & 0.452$\pm$0.065  & 0.563$\pm$0.071 & 0.614$\pm$0.061 \\ \hline
\end{tabular}}
\label{tab: performance on other datasets}
\vspace{-1em}
\end{table*}

\subsubsection*{Predictors}
Predictors are defined as the error samples obtained by computing a distance over a descriptor. We define different distance functions for different descriptors. We use the Euclidean distance to measure the point-to-point distances between query point pairs under the new basis 
\begin{equation}
r_{\alpha} = \sqrt{\textstyle \sum_{m} {\bm{d_{1}}}^2},
\label{eq:Euclidean}
\end{equation}
where $\bm{d_{1}}$ is the difference between two points. 
We use the absolute value to measure the point-to-plane distance, as 
\begin{equation}
r_{\beta} = |\bm{d_{2}}|,
\label{eq:Projected}
\end{equation}
where $\bm{d_{2}}$ indicates the projected distance between a point and the reference axes.
We use the following definition of relative difference for the covariance features
\begin{equation}
r_{\gamma} = \frac{|\bm{q}^{\mathcal{A}}\odot \bm{q}^{\mathcal{B}} -\bm{Q}|}{\bm{q}^{\mathcal{A}}\odot \bm{q}^{\mathcal{B}}},
\label{eq:relative covaraince distance}
\end{equation}
where $\lbrace \bm{q}^{\mathcal{F}} = \bm{\lambda}^{\mathcal{F}}, \mathbf{Q} = \bm{\Sigma}\rbrace$ and 
$\lbrace \bm{q}^{\mathcal{F}} = \tilde{\bm{s}}^{\mathcal{F}}, \mathbf{Q} = \tilde{\bm{\Sigma}}\rbrace$, for geometry and texture attributes, respectively, $\odot$ is for element-wise product. We use the relative difference formula \cite{9106005}, for the remaining descriptors 
\begin{equation}
r_{\delta}= \frac{|\phi^{\mathcal{A}} - \phi^{\mathcal{B}}|} {\left| \phi^{\mathcal{A}} \right| + \left| \phi^{\mathcal{B}} \right|  + \varepsilon},
\label{eq:relDiff}
\end{equation}
where $\varepsilon$ is a small constant to avoid undefined operations. Finally, the definitions of parallelity and angular similarity descriptors incorporate a distance function. For notational purposes only, we define distances $r_{\rho}$ and $r_{\theta}$ to be identical to the definitions of $\mathscr{P}_{m}$ and $\theta_{m}$, respectively.
Table~\ref{tbl:descript} enlists distance function(s) used per descriptor.

\subsubsection*{Features} Features are defined by pooling over predictor values. 
Specifically, 

predictors $\psi_{i,j,k}$ are obtained per point $\mathbf{p}_i$, descriptor $j$, and distance function $r_k$, 
$k\in \{\alpha, \beta, \gamma, \delta, \rho, \theta\}$. 
This is done for all descriptors $j$ in Table~\ref{tbl:descript}, using the corresponding distances $r_k$.
Through pooling, we obtain a feature $f_{j,k}$ for every predictor:

\begin{equation}
f_{j,k} = \frac{1}{|\mathcal{A}|} \textstyle \sum_{i = 1}^{|\mathcal{A}|} \psi_{i,j,k}.
\label{eq:pool}
\end{equation}

\subsection{Quality Regression}
To obtain a quality score that is well-aligned with the HVS, RFE is used to select the most relevant predictor set among all the proposed predictors. RFE \cite{guyon2002gene} improves model accuracy, and efficiency, and reduces overfitting. 
We then use the random forest algorithm to regress the selected predictors to the final quality score.

\section{EXPERIMENTAL RESULTS}
In this section, we report evaluation results of the proposed PointPCA+ metric under three public datasets. Moreover, we report the performance achieved in the ICIP 2023 Point Cloud Visual Quality Assessment (PCVQA) grand challenge\footnote{\url{https://sites.google.com/view/icip2023-pcvqa-grand-challenge}}.
Specifically, the challenge consists of 5 tracks, which correspond to different use cases in which quality metrics are typically used. 
We participated in Track$\#$1 FR broad-range quality estimation, Track$\#$3 FR high-range quality estimation, and Track$\#$5 Intra-reference quality estimation. 

\subsection{Setup}
\subsubsection*{Datasets}
Three publicly available datasets were recruited for performance evaluation, namely, M-PCCD, SJTU, and WPC.
The M-PCCD \cite{alexiou2019comprehensive} consists of 8 point clouds whose geometry and color are encoded using V-PCC and G-PCC variants, resulting in 232 distorted stimuli. 
The SJTU \cite{yang2020predicting} includes 9 reference point clouds with each point cloud corrupted by seven types of distortions under six levels, generating $378$
distorted stimuli. 
The WPC \cite{9756929} contains 20 reference point clouds with each point cloud degraded under five types of distortions and different levels, leading to $740$ 
distorted stimuli.
The Broad Quality Assessment of Static Point Clouds (BASICS) \cite{ak2023basics} is used in the ICIP 2023 PCVQA grand challenge, and comprises 75 point clouds from 3 different semantic categories. Each point cloud is compressed with 4 different algorithms at varying compression levels, resulting in 1494 processed point clouds. 

\subsubsection*{Evaluation metrics}
We evaluate the performance with four standard criteria: Pearson Linear Correlation Coefficient (PLCC), Spearman Rank Order Correlation Coefficient (SROCC), Difference/Similar Analysis quantified by Area Under the Curve (AUC), and Better/Worse Analysis quantified by Correct Classification percentage (CC) \cite{7498936}. No function is adopted for score mapping. Finally, we report the Runtime Complexity (RC).

\subsubsection*{Implementation details} We use RFE to select the best feature set among all the predictors, with the best SROCC performance on the training set. In the inference stage, the default configuration of scikit-learn (version 1.2.2) in Python is used. 
Regarding the neighborhood size for the computation of descriptors, $K=81$ is chosen considering complexity and performance, after experimenting with $K\in {\{9,25,49,81,121\}}$.

\subsection{Overall performance comparison}

\subsubsection*{Performance evaluation on M-PCCD, SJTU and WPC}
We compare PointPCA+ with existing state-of-the-art FR quality metrics, the results
are shown in TABLE \ref{tab: performance on other datasets}. The best performance among these metrics is highlighted in boldface, with the second best underlined. We used an 80\%-20\% split for training and testing for each dataset. Then, the average and the standard deviation of SROCC index computed across all testing splits of each dataset, are reported. Specifically, for M-PCCD, SJTU, and WPC, we have 28, 36, and 4845 splits respectively. PCA-based metrics are competitive with the highest PLCC/SROCC on the three datasets.

\subsubsection*{Performance evaluation on BASICS}
We split BASICS into training-validation-test with 60\%-20\%-20\% following the rules from the PCVQA grand challenge \cite{ICIP2023}. Table \ref{tab: performance on basics}  shows the official evaluation results of Track$\#$1. The PointPCA+ secured a third-place ranking in Track$\#$3 and Track$\#$5,  with \{PLCC, SROCC, AUC, CC\}$=$  \{$0.479, 0.603, 0.625, 0.886$\} and \{AUC, CC\} $=$ \{$0.811, 0.938$\}, respectively.

\begin{table}[htbp]
\caption{Top 4 performance comparison on the official PCVQA grand challenge test set, evaluated by the challenge organizers. Best in bold and second best underlined. }
\centering
\vspace{-0.5em}
\renewcommand{\arraystretch}{1.1}
\scalebox{0.9}{
\begin{tabular}{c|c|c|c|c|c}
\hline
Submission   & PLCC            & SROCC           & AUC             & CC & RC(s)   \\ \hline
 KDDIUSCJoint & \textbf{0.917} & \textbf{0.875}    & \textbf{0.888} & \textbf{0.970} &\underline{42.8}       \\
 CWI\_DIS     & \underline{0.909}          & \underline{0.874} & \underline{0.871}         & \underline{0.961}   &1000       \\

 SJTU MMLAB   & 0.896     & 0.871   &  0.832    &0.955 &\textbf{8.60} \\
 SlowHand     & 0.825      & 0.791  & 0.805     & 0.924 &130.47	    \\\hline
\end{tabular}}
\label{tab: performance on basics}

\vspace{-0.5em}
\end{table}

\section{Conclusion}
This paper proposes a PCA-based PCQA metric, namely PointPCA+, which relies on an enriched set of lower complexity descriptors with respect to its PointPCA predecessor.
After a pre-processing step, features are extracted from both geometric and textural domains. A subset of features is selected to enhance the stability of the model, and a learning-based feature fusion based on ensemble learning is applied to the feature subset, to provide a total quality score for a distorted point cloud.
Our experimental results demonstrate that PointPCA+ outperforms the majority of existing PCQA metrics, reaching second place in Track$\#1$ of the ICIP 2023 PCVQA grand challenge. 
Future work will focus on further reducing computational complexity and incorporating global descriptors to more effectively handle general distortion.

\section{Acknowledgements}
This work was supported through the NWO WISE grant and the European Commission Horizon Europe program, under the grant agreement 101070109, \textit{TRANSMIXR} \url{https://transmixr.eu/}. Funded by the European Union.

\bibliographystyle{IEEEbib}
\bibliography{refs}

\begin{thebibliography}{10}

\bibitem{ALEXIOU2023501}
Evangelos Alexiou, Yana Nehmé, Emin Zerman, Irene Viola, Guillaume Lavoué, Ali Ak, Aljosa Smolic, Patrick {Le Callet}, and Pablo Cesar,
\newblock ``Chapter 18 - subjective and objective quality assessment for volumetric video,''
\newblock in {\em Immersive Video Technologies}, Giuseppe Valenzise, Martin Alain, Emin Zerman, and Cagri Ozcinar, Eds., pp. 501--552. Academic Press, 2023.

\bibitem{8486512}
Evangelos Alexiou and Touradj Ebrahimi,
\newblock ``Point cloud quality assessment metric based on angular similarity,''
\newblock in {\em 2018 IEEE International Conference on Multimedia and Expo (ICME)}, 2018, pp. 1--6.

\bibitem{8743277}
Evangelos Alexiou and Touradj Ebrahimi,
\newblock ``Exploiting user interactivity in quality assessment of point cloud imaging,''
\newblock in {\em 2019 Eleventh International Conference on Quality of Multimedia Experience (QoMEX)}, 2019, pp. 1--6.

\bibitem{9756929}
Qi~Liu, Honglei Su, Zhengfang Duanmu, Wentao Liu, and Zhou Wang,
\newblock ``Perceptual quality assessment of colored 3d point clouds,''
\newblock {\em IEEE Transactions on Visualization and Computer Graphics}, pp. 1--1, 2022.

\bibitem{9123089}
Irene Viola, Shishir Subramanyam, and Pablo Cesar,
\newblock ``A color-based objective quality metric for point cloud contents,''
\newblock in {\em 2020 Twelfth International Conference on Quality of Multimedia Experience (QoMEX)}, 2020, pp. 1--6.

\bibitem{9106005}
Evangelos Alexiou and Touradj Ebrahimi,
\newblock ``Towards a point cloud structural similarity metric,''
\newblock in {\em 2020 IEEE International Conference on Multimedia Expo Workshops (ICMEW)}, 2020, pp. 1--6.

\bibitem{9306905}
Qi~Yang, Zhan Ma, Yiling Xu, Zhu Li, and Jun Sun,
\newblock ``Inferring point cloud quality via graph similarity,''
\newblock {\em IEEE Transactions on Pattern Analysis and Machine Intelligence}, vol. 44, no. 6, pp. 3015--3029, 2022.

\bibitem{9123147}
Gabriel Meynet, Yana Nehmé, Julie Digne, and Guillaume Lavoué,
\newblock ``{PCQM}: A full-reference quality metric for colored 3d point clouds,''
\newblock in {\em 2020 Twelfth International Conference on Quality of Multimedia Experience (QoMEX)}, 2020, pp. 1--6.

\bibitem{diniz2022point}
Rafael Diniz, Pedro~Garcia Freitas, and Mylene~CQ Farias,
\newblock ``Point cloud quality assessment based on geometry-aware texture descriptors,''
\newblock {\em Computers \& Graphics}, 2022.

\bibitem{liu2023point}
Yipeng Liu, Qi~Yang, Yiling Xu, and Le~Yang,
\newblock ``Point cloud quality assessment: Dataset construction and learning-based no-reference metric,''
\newblock {\em ACM Transactions on Multimedia Computing, Communications and Applications}, vol. 19, no. 2s, pp. 1--26, 2023.

\bibitem{liu2022progressive}
Qi~Liu, Yiyun Liu, Honglei Su, Hui Yuan, and Raouf Hamzaoui,
\newblock ``Progressive knowledge transfer based on human visual perception mechanism for perceptual quality assessment of point clouds,''
\newblock {\em arXiv preprint arXiv:2211.16646}, 2022.

\bibitem{zhang2022mm}
Zicheng Zhang, Wei Sun, Xiongkuo Min, Quan Zhou, Jun He, Qiyuan Wang, and Guangtao Zhai,
\newblock ``{MM-PCQA}: Multi-modal learning for no-reference point cloud quality assessment,''
\newblock {\em arXiv preprint arXiv:2209.00244}, 2022.

\bibitem{alexiou2021pointpca}
Evangelos Alexiou, Xuemei Zhou, Irene Viola, and Pablo Cesar,
\newblock ``Pointpca: Point cloud objective quality assessment using pca-based descriptors,''
\newblock {\em arXiv preprint arXiv:2111.12663}, 2021.

\bibitem{8296925}
Dong Tian, Hideaki Ochimizu, Chen Feng, Robert Cohen, and Anthony Vetro,
\newblock ``Geometric distortion metrics for point cloud compression,''
\newblock in {\em 2017 IEEE International Conference on Image Processing (ICIP)}, 2017, pp. 3460--3464.

\bibitem{ITURBT7096}
{ITU-R BT.709-6},
\newblock ``{Parameter values for the {HDTV} standards for production and international programme exchange},'' International Telecommunication Unionn, Jun. 2015.

\bibitem{9450013}
Rafael Diniz, Pedro~Garcia Freitas, and Mylène C.~Q. Farias,
\newblock ``Color and geometry texture descriptors for point-cloud quality assessment,''
\newblock {\em IEEE Signal Processing Letters}, vol. 28, pp. 1150--1154, 2021.

\bibitem{guyon2002gene}
Isabelle Guyon, Jason Weston, Stephen Barnhill, and Vladimir Vapnik,
\newblock ``Gene selection for cancer classification using support vector machines,''
\newblock {\em Machine learning}, vol. 46, pp. 389--422, 2002.

\bibitem{alexiou2019comprehensive}
Evangelos Alexiou, Irene Viola, Tom{\'a}s~M Borges, Tiago~A Fonseca, Ricardo~L De~Queiroz, and Touradj Ebrahimi,
\newblock ``A comprehensive study of the rate-distortion performance in mpeg point cloud compression,''
\newblock {\em APSIPA Transactions on Signal and Information Processing}, vol. 8, pp. e27, 2019.

\bibitem{yang2020predicting}
Qi~Yang, Hao Chen, Zhan Ma, Yiling Xu, Rongjun Tang, and Jun Sun,
\newblock ``Predicting the perceptual quality of point cloud: A 3d-to-2d projection-based exploration,''
\newblock {\em IEEE Transactions on Multimedia}, vol. 23, pp. 3877--3891, 2020.

\bibitem{ak2023basics}
Ali Ak, Emin Zerman, Maurice Quach, Aladine Chetouani, Aljosa Smolic, Giuseppe Valenzise, and Patrick~Le Callet,
\newblock ``Basics: Broad quality assessment of static point clouds in compression scenarios,''
\newblock {\em arXiv preprint arXiv:2302.04796}, 2023.

\bibitem{7498936}
Lukáš Krasula, Karel Fliegel, Patrick Le~Callet, and Miloš Klíma,
\newblock ``On the accuracy of objective image and video quality models: New methodology for performance evaluation,''
\newblock in {\em 2016 Eighth International Conference on Quality of Multimedia Experience (QoMEX)}, 2016, pp. 1--6.

\bibitem{ICIP2023}
Aladine Chetouani, Giuseppe Valenzise, Ali Ak, Emin Zerman, Maurice Quach, Marouane Tliba, Mohamed~Amine Kerkouri, and Patrick~Le Callet,
\newblock ``{ICIP} 2023 - point cloud visual quality assessment grand challenge,''
\newblock 2023.

\end{thebibliography}

\end{document}